\def\year{2020}\relax
\documentclass[letterpaper]{article}
\usepackage{aaai20}
\usepackage{times}
\usepackage{helvet}
\usepackage{courier}
\usepackage[hyphens]{url}
\usepackage{graphicx}
\usepackage{subfigure}

\usepackage[ruled, vlined, linesnumbered, commentsnumbered]{algorithm2e}
\DontPrintSemicolon
\SetKwComment{tch}{\# }{}
\SetKwBlock{Procedure}{}{}
\SetKwBlock{Function}{function}{}
\newcommand{\Comment}[1]{\tch*[h]{\small #1}}

\usepackage{amsthm}
\usepackage{amsmath}
\usepackage{amsfonts}

\theoremstyle{plain}

\theoremstyle{definition}

\newcommand{\citet}[1]{\citeauthor{#1} (\citeyear{#1})}

\makeatletter
\renewcommand{\paragraph}{%
  \@startsection{paragraph}{4}%
  {\z@}{1.25ex \@plus 1ex \@minus .2ex}{-1em}
  {\normalfont\normalsize\bfseries}%
}
\makeatother

\nocopyright

\title{Multi-Agent Deep Reinforcement Learning with Adaptive Policies}
\author{Yixiang Wang \and Feng Wu \\
School of Computer Science and Technology \\
University of Science and Technology of China \\
{\tt yixiangw@mail.ustc.edu.cn}, {\tt wufeng02@ustc.edu.cn}}

\begin{document}

\maketitle

\begin{abstract}
We propose a novel approach to address one aspect of the non-stationarity problem in multi-agent reinforcement learning (RL), where the other agents may alter their policies due to environment changes during execution. This violates the Markov assumption that governs most single-agent RL methods and is one of the key challenges in multi-agent RL. To tackle this, we propose to train multiple policies for each agent and postpone the selection of the best policy at execution time. Specifically, we model the environment non-stationarity with a finite set of scenarios and train policies fitting each scenario. In addition to multiple policies, each agent also learns a policy predictor to determine which policy is the best with its local information. By doing so, each agent is able to adapt its policy when the environment changes and consequentially the other agents alter their policies during execution. We empirically evaluated our method on a variety of common benchmark problems proposed for multi-agent deep RL in the literature. Our experimental results show that the agents trained by our algorithm have better adaptiveness in changing environments and outperform the state-of-the-art methods in all the tested environments.
\end{abstract}

\section{Introduction}

The development of modern deep learning has made reinforcement learning (RL) more powerful to solve complex decision problems. This leads to success in many real-world applications, such as Atari games \cite{mnih2015human}, playing Go \cite{silver2016mastering} and robotics control \cite{levine2016end}. Recently, there is growing focus on applying deep RL techniques to multi-agent systems. Many promising approaches for multi-agent deep RL have been proposed to solve a variety of multi-agent problems, such as traffic control \cite{wu2017flow}, multi-player games (e.g., StarCraft, Dota 2), and multi-robot systems \cite{long2018towards}.

Despite the recent success of deep RL in single-agent domains, there are additional challenges in multi-agent RL. One major challenge is the {\em non-stationarity} of multi-agent environment caused by agents that change their policies during the training and testing procedures. Specifically, at the training time, each agent's policy is changing simultaneously and therefore the environment becomes non-stationary from the perspective of any individual agent. To handle this issue, {\em multi-agent deep deterministic policy gradient} (MADDPG) \cite{lowe2017multi} proposed to utilized a {\em centralized critic} with {\em decentralized actors} in the actor-critic learning framework. Since the centralized Q-function of each agent is conditioned on the actions of all the other agents, each agent can perceive the learning environment as stationary even when the policies of the other agents change.

Although using a centralized critic stabilizes training, the learned policy of each agent can still be brittle and sensitive to its training environment and partners. It has been observed that the performance of the learned policies can be drastically worse when some agents alter their policies during execution \cite{lazaridou2016multi}. To improve the robustness of the learned policies, {\em minimax multi-agent deep deterministic policy gradient} (M3DDPG) \cite{li2019robust} --- a {\em minimax} extension of MADDPG --- proposed to update policies considering the worst-case situation by assuming that all the other agents acts adversarially. This minimax optimization is useful to learn robust policies in very competitive domains but can be too {\em pessimistic} in mixed competitive and cooperative or fully cooperative problems as shown later in our experiments.

In this paper, we consider one aspect of the non-stationarity issue in multi-agent RL, where the other agents may alter their policies as a result of changes in some environmental factors. This frequently happens in real-world activities. For example, in a soccer game, a heavy rain or high temperature usually causes the teams to change their strategies against each other. Take disaster response as another example. First responders often need to constantly adjust their plan in order to complete their tasks in the highly dynamic and danger environment. Therefore, it is often desirable for the agents to learn policies that can adapt with changes of the environment and the other agents' policies.

Against this background, we propose {\em policy adaptive multi-agent deep deterministic policy gradient} (PAMADDPG) --- a novel approach based on MADDPG --- to learn adaptive policies for non-stationary environments. Specifically, it
learns multiple policies for each agent and postpone the selection of the best policy at execution time. By doing so, each agent is able to adapt its policy when the environment changes. Specifically, we model the non-stationary environment by a finite set of known scenarios, where each scenario captures possible changing factors of the environment (e.g., weather, temperature, wind, etc. in soccer). For each scenario, a policy is learned by each agent to perform well in that specific scenario. Together with multiple policies for each agent, we also train a policy predictor to predict the best policy using the agent's local information. At execution time, each agent first selects a policy based on the policy predictor and then choose an action according to the selected policy. We evaluated our PAMADDPG algorithm on three common benchmark environments and compared it with MADDPG and M3DDPG. Our experimental results show that PAMADDPG outperforms both MADDPG and M3DDPG in all the tested environments.

The rest of the paper is organized as follows. We first briefly review the related work about handling non-stationary in multi-agent deep RL. Then, we describe the background on the Markov game and the MADDPG method, which are building blocks of our algorithm. Next, we propose our PAMADDPG algorithm to learn multiple policies and policy predictors. After that, we present the experiments with environments, setup, and results. Finally, we conclude the paper with possible future work.

\section{Related Work}

In recent years, various approaches \cite{papoudakis2019dealing} have been proposed to tackle different aspects of non-stationarity in multi-agent deep RL. We sample a few related work about multi-agent deep RL as listed below.

\paragraph{Centralized critic.}
Using the centralized critic techniques, \citet{lowe2017multi} proposed MADDPG for multi-agent RL using a centralized critic and a decentralized actor, where the training of each agent is conditioned on the observation and action of all the other agents so the agent can perceive the environment as stationary. \citet{li2019robust} extended MADDPG and proposed M3DDPG using minimax Q-learning in the critic to exhibit robustness against different adversaries with altered policies. \citet{foerster2018counterfactual} proposed COMA using also a centralized critic with the counterfactual advantage estimation to address the credit assignment problem --- another key challenge in multi-agent RL.

\paragraph{Decentralized learning.}
A useful decentralized learning technique to handle non-stationarity is self-play. Recent self-play approaches store the neural network parameters at different points during learning. By doing so, self-play managed to train policies that can generalize well in environments like Go \cite{silver2017mastering} and complex locomotion tasks \cite{bansal2017emergent}. Another technique \cite{foerster2017stabilising} is by stabilizing experience replay using importance sampling corrections to adjust the weight of previous experience to the current environment dynamics.

\paragraph{Opponent modeling.}
By modeling the opponent, \citet{he2016opponent} developed a second separate network to encode the opponent's behaviour. The combination of the two networks is done either by concatenating their hidden states or by the use of a mixture of experts. In contrast, \citet{raileanu2018modeling} proposed an actor-critic method using the same policy network for estimating the goals of the other agents. \citet{foerster2018learning} proposed a modification of the optimization function to incorporate the learning procedure of the opponents in the training of agents.

\paragraph{Meta-learning.}
 By extending meta-learning approaches for single-agent RL such as model agnostic meta-learning \cite{finn2017model} to handle non-stationarity in multi-agent domains, \citet{al2017continuous} proposed an optimization method to search for initial neural network parameters that can quickly adapt to non-stationarity, by explicitly optimizing the initial model parameters based on their expected performance after learning. This was tested in {\em iterated adaptation games}, where an agent repeatedly play against the same opponent while only allowed to learn in between each game.

\paragraph{Communication.}
In this direction, \citet{foerster2016learning} proposed the deep distributed recurrent Q-networks, where all the agents share the same hidden layers and learn to communicate to solve riddles. \citet{sukhbaatar2016learning} proposed the CommNet architecture, where the input to each hidden layer is the previous layer and a communication message. \citet{singh2018learning} proposed the individualized controlled continuous communication model, which is an extension of CommNet in competitive setting. \citet{foerster2016learningto} proposed reinforced inter-agent learning with two Q-networks for each agents where the first network outputs an action and the second a communication message.

As briefly reviewed above, most of the existing work focus on handling non-stationarity mainly during training procedure. Although meta-learning approaches can learn to adapt agents' policies between different game, it requires to {\em repeatedly} play iterated adaptation games. In contrast, we build our algorithm on top of MADDPG to address the non-stationarity problem in general multi-agent RL at execution time. Additionally, we do not assume explicit communication among the agents during execution as in MADDPG.

A complete survey about recent efforts of dealing non-stationarity in multi-agent RL can be found in \cite{hernandez2017survey,papoudakis2019dealing}.

\section{Background}

In this section, we introduce our problem settings and some
basic algorithms on which our approach is based.

\subsection{Partially Observable Markov Games}

In this work, we consider a {\em partially observable Markov games} \cite{littman1994markov} with $N$ agents, defined by: a set of states $\mathcal{S}$ describing the possible configurations of all agents, a set of actions $\mathcal{A}_1,\ldots,\mathcal{A}_N$ and a set of observations $\mathcal{O}_1,\ldots,\mathcal{O}_N$ for each agent. To choose actions, each agent $i$ uses a stochastic policy $\mu_{\theta_i} : \mathcal{O}_i \times \mathcal{A}_i \mapsto [0,1]$, which produces the next state according to the state transition function $\mathcal{T} : \mathcal{S} \times \mathcal{A}_1 \times \ldots \times \mathcal{A}_N \mapsto \mathcal{S}$.

At each time step, each agent $i$ obtains rewards as a function of the state and agent's action $r_i : \mathcal{S} \times \mathcal{A}_i \mapsto \mathbb{R}$, and receives a local observation correlated with the state $\mathbf{o}_i : \mathcal{S} \mapsto \mathcal{O}_i$. The initial states are determined by a state distribution $\rho : \mathcal{S} \mapsto [0,1]$. Each agent $i$ aims to maximize its own total expected return: $R_i = \sum_{t=0}^T \gamma^t r^t_i$, where $\gamma \in (0, 1]$ is a discount factor and $T$ is the time horizon.

Here, we assume that the state transition function $\mathcal{T}$ is unknown and therefore consider to learn the policies $\mu_{\theta_i}$ for each agent $i$ using multi-agent {\em reinforcement learning} (RL) methods. Note that each agent must choose an action based on its own policy and local observation during execution.

\subsection{Multi-Agent Deep Deterministic Policy Gradient}

Policy gradient methods are a popular choice for a variety of RL tasks. The main idea is to directly adjust the parameters $\theta$ of the policy in order to maximize the objective $J(\theta) = \mathbb{E}_{s \sim p^{\mu}, a \sim \mu_\theta}[R(s, a)]$ by taking steps in the direction of $\nabla_\theta J(\theta)$, i.e., the gradient of the policy written as:
\begin{equation}
\nabla_\theta J(\theta) = \mathbb{E}_{s \sim p^\mu, a \sim \mu_\theta} [\nabla_\theta \log \mu_\theta(a|s) Q^\mu (s,a)]
\end{equation}
where $p^\mu$ is the state distribution and $Q^\mu$ is the Q-function.

The policy gradient framework has been extended to deterministic policies $\mu_\theta: \mathcal{S} \mapsto \mathcal{A}$. In particular, under certain conditions the gradient of the objective $J(\theta) = \mathbb{E}_{s \sim p^{\mu}}[R(s,a)]$ can be written as:
\begin{equation}
\nabla_\theta J(\theta) = \mathbb{E}_{s \sim \mathcal{D}} [\nabla_\theta \mu_\theta(a|s) \nabla_a Q^{\mu} (s,a)|_{a=\mu_\theta (s)}]
\end{equation}
Since the {\em deterministic policy gradient} (DPG) \cite{silver2014deterministic} relies on $\nabla_a Q^{\mu} (s,a)$, it requires that the action space $\mathcal{A}$ (and thus the policy $\mu$) be continuous. {\em Deep deterministic policy gradient} (DDPG) \cite{lillicrap2015continuous} is a variant of DPG where the policy $\mu$ and critic $Q^{\mu}$ are approximated with deep neural networks. DDPG is an off-policy algorithm, and samples trajectories from a replay buffer of experiences that are stored throughout training. It also makes use of a target network, as in DQN \cite{mnih2015human}.

{\em Multi-agent DDPG} (MADDPG) \cite{lowe2017multi} extends the DDPG method to multi-agent domains. The main idea behind MADDPG is to consider action policies of other agents. The environment is stationary even as the policies change, since $P(s'|s,a_1,\ldots,a_N,\pi_1,\ldots,\pi_N) = P(s'|s,a_1,\ldots,a_N) = P(s'|s,a_1,\ldots,a_N,\pi'_1,\ldots,\pi'_N)$ for any $\pi_i \neq \pi_i'$. The gradient can be written as:
\begin{equation}
\begin{split}
\nabla_{\theta_i} J(\mu_i) =
\mathbb{E}_{\mathbf{x},a \sim \mathcal{D}}[ & \nabla_{\theta_i} \mu_i(a_i|o_i) \nabla_{a_i} \\
& Q^{\mu}_i (\mathbf{x}, a_1, \ldots, a_N) |_{a_i=\mu_i (o_i)}]
\end{split}
\end{equation}
where $Q^{\mu}_i (\mathbf{x}, a_1, ..., a_N)$ is a \textit{centralized action-value function} that takes as input the actions of all agents, $a_1,\ldots, a_N$, in addition to the state information $\mathbf{x}$, and outputs the Q-value for agent $i$. Here, $Q^{\mu}_i$ can be updated as:
\begin{equation}
\begin{split}
\mathcal{L}(\theta_i) = \mathbb{E}_{\mathbf{x}, a, r, \mathbf{x}'} [(Q^{\mu}_i(\mathbf{x}, a_1, \dots, a_N) - y)^2], \\
y = r_i + \gamma Q^{\mu'}_i (\mathbf{x}', a'_1, \dots, a'_N)|_{a'_j = \mu'_j(o_j)}
\end{split}
\end{equation}
where $(\mathbf{x}, a, r, \mathbf{x}')$ is sampled from the experience replay buffer $\mathcal{D}$, recoding experiences of all agents.

\subsection{Dealing Non-Stationarity in MADDPG}

As aforementioned, one of the key challenges in multi-agent RL is the environment non-stationarity. This non-stationarity stems from breaking the Markov assumption that governs most single-agent RL algorithms. Since the transitions and rewards depend on actions of all agents, whose decision policies keep changing in the learning process, each agent can enter an endless cycle of adapting to other agents. Although using a centralized critic stabilizes training in MADDPG, the learned policies can still be brittle and sensitive to changes of the other agents's policies.

To obtain policies that are more robust to changes in the policy of other agents, MADDPG proposes to first train a collection of $K$ different sub-policies and then maximizing the ensemble objective $\max_{\theta_i} J(\theta_i)$ as:
\begin{equation}
\begin{split}
  J(\theta_i) & = \mathbb{E}_{k \sim \mathrm{uniform}(1, K), s \sim p^{\mu}, a\sim \mu^{(k)}} [ R_i(s, a) ] \\
  & = \mathbb{E}_{k, s} \left[ \sum_{t=0}^T \gamma^t r_i( s^t, a^t_1, \dots, a^t_N) \Big|_{a^t_i = \mu^{(k)}_i(o^t_i)} \right] \\
  & = \mathbb{E}_{s} \left[ \frac{1}{K} \sum_{k=1}^K Q^{\mu}_{i}(s, a_1, \dots, a_N) \Big|_{a_i = \mu^{(k)}_i(o_i)} \right]
\end{split}
\end{equation}
where $\mu^{(k)}_i$ is the $k$-th sub-policies of agent $i$. By training agents with an ensemble of policies, the agents require interaction with a variety of the other agents' policies. Intuitively, this is useful to avoid converging to local optima of the agents' policies. However, the ensemble objective only considers the {\em average} performance of agents' policies training by {\em uniformly} sampling the policies of the other agents.

Alternatively, M3DDPG \cite{li2019robust} --- a variation of MADDPG --- proposes to update policies considering the worst situation for the purpose of learning robust policies. During training, it optimizes the policy of each agent $i$ under the assumption that all other agents acts adversarially, which yields the minimax objective $\max_{\theta_i} J(\theta_i)$ as:
\begin{equation}
\begin{split}
  J(\theta_i) & = \min_{a_{j\neq i}} \mathbb{E}_{s \sim p^{\mu}, a_i \sim \mu_i} [ R_i(s, a) ] \\
  & = \min_{a^t_{j\neq i}} \mathbb{E}_{s} \left[ \sum_{t=0}^T \gamma^t r_i(s^t, a^t_1, \dots, a^t_N) \Big|_{a^t_i=\mu_i(o^t_i)} \right] \\
  & = \mathbb{E}_{s} \left[ \min_{a_{j\neq i}} Q^{\mu}_{M, i}(s, a_1, \dots, a_N) \Big|_{a_i=\mu_i(o_i)} \right]
\end{split}
\end{equation}
where $Q^{\mu}_{M, i}(s, a_1, \dots, a_N)$ is the modified Q function representing the current reward of executing $a_1, \dots, a_N$ in $s$ plus the discounted worst case future return starting from $s$. With the minimax objective, the training environment of each agent becomes stationary because the behavior of all the other agents only depends on $-r_i$, i.e., the negative reward of agent $i$ itself. However, this adversarial assumption could be too {\em pessimistic} if the game among the agents is not zero-sum or even is cooperative.

Ideally, the well trained agents should be able to {\em adapt} their policies with the changes in the environment. This motivated the development of our algorithm introduced next.

\section{Policy Adaptive Multi-Agent Deep Deterministic Policy Gradient}

\begin{algorithm}[t]
  \caption{Training and execution for PAMADDPG}
  \label{alg:main}
  \Comment{At training time:} \\
  $\forall i: \Pi_i \gets \emptyset, \phi_i \gets$ initialize the predictor parameters \\
  \ForEach{scenario $c\in \mathcal{C}$}{
    $\forall i: \Pi_i \gets$ learn and add a set of policies for agent $i$ \\
    $\forall i: \phi_i \gets$ learn and update the predictor for agent $i$
  }
  \Comment{At execution time:} \\
  $\forall i: h^0_i \gets \emptyset$ \\
  \For{time step $t = 1$ \KwTo $T$}{
    \For{agent $i = 1$ \KwTo $N$}{
        $o_i^t \gets$ receive a local observation for agent $i$ \\
        $\mu_i \gets$ select a policy from $\Pi_i$ by $\phi_i(o_i^t, h_i^{t-1})$ \\
        $a_i^t \gets$ select an action by $\mu_{\theta_i}(o_i^t)$ \\
        $h_i^t \gets$ append $o_i^t$ to $h_i^{t-1}$
    }
    Execute actions $\langle a_1^t, \dots, a_N^t \rangle$ to the environment \\
    Collect rewards $\langle r_1^t, \dots, r_N^t \rangle$ from the environment
  }
  \Return $\forall i: R_i = \sum_{t=0}^T \gamma^t r_i^t$
\end{algorithm}

In this section, we propose {\em policy adaptive multi-agent deep deterministic policy gradient} (PAMADDPG), which is based on MADDPG, to deal with environment non-stationarity in multi-agent RL. As in MADDPG, our algorithm operate under the framework of {\em centralized training} with {\em decentralized execution}. Thus, we allow the agents to share extra information for training, as long as this information is not used at execution time. We assume that the learned policies can only use local information and there is no explicit communication among agents during execution. Specifically, our algorithm is an extension of actor-critic policy gradient methods with multiple decentralized actors and one centralized critic, where the critic is augmented with extra information on the policies of the other agents.

In this work, we consider a setting where agents are trained and executed in an environment that can categorized into a finite set of scenarios. These scenarios are known during training. However, at execution time, agents have no prior knowledge about which scenario they will locate in. Therefore, the agents must act adaptively during execution. Note that the scenarios cannot be modeled as state variables because we make no assumption about the initial distribution and transition probabilities of scenarios, which can be any probabilities in our setting. Intuitively, a scenario in our setting models a collection of environmental factors that can cause the agents to alter their policies.

Let $\mathcal{C}$ denote a finite set of scenarios for the agents. Here, each scenario $c\in \mathcal{C}$ can be modeled by a partially observable Markov game as aforementioned. We assume that all the scenarios in $\mathcal{C}$ have identical state space and the same action and observation space for all the agents. Particularly, each scenario $c\in \mathcal{C}$ may have different state transition function $\mathcal{T}^c$ and different reward function $r^c_i$ for each agent $i$, so that agents in different scenarios may require different policies. Formally, we define a scenario $c \in \mathcal{C}$ as a tuple: $\langle \mathcal{S}, \{ \mathcal{A}_i \}, \{ \mathcal{O}_i \}, \mathcal{T}^c, \{ r^c_i \} \rangle$ with notations in Markov games.

As aforementioned, to be able to adapt in different scenarios, we propose to train multiple policies for each agent and postpone the selection of its policy at execution time. In addition to multiple policies for each agent, we also train a policy predictor that can be used by the agent to determine the best policy during execution. Given this, the agent is able to adapt its policy when the environment changes. As summarized in Algorithm \ref{alg:main}, PAMADDPG consists of two main procedures: 1) learning multiple policies and 2) learning policy predictors, which will be described in details next.

\subsection{Learning Multiple Policies}

We can extend the actor-critic policy gradient method as described in MADDPG to work with each scenario. Specifically, given a scenario $c \in \mathcal{C}$, the gradient for policy $\mu_i^c$ with respect to parameters $\theta_i^c$ can be written as:
\begin{equation}
\begin{split}
    \nabla_{\theta_i^c} J(\mu_i^c) = \mathbb{E}_{\mathbf{x}, a \sim \mathcal{D}^c}[ \nabla_{\theta_i^c} \mu_i^c(a_i|o_i) \nabla_{a_i} \\ Q^{\mu, c}_i (\mathbf{x}, a_1, \dots, a_N) \big|_{a_i = \mu_i^c (o_i)}]
\end{split}
\end{equation}
where $\mathcal{D}^c$ is the experience replay buffer recording experiences with tuples $(\mathbf{x}, a_1, \dots, a_N, r_1^c, \dots, r_N^c, \mathbf{x}')$ of all agents at the scenario $c$ and $\mathbf{x} = (o_1, \dots, o_N)$. Here, the centralized action-value function $Q^{\mu, c}_i$ is updated as:
\begin{equation}
\begin{split}
\mathcal{L}(\theta_i^c) = \mathbb{E}_{\mathbf{x}, a, r, \mathbf{x}'}[(Q^{\mu,c}_i(\mathbf{x}, a_1, \dots, a_N) - y)^2] \\
y = r_i + \gamma\, Q^{\mu', c}_i(\mathbf{x}', a_1', \dots, a_N') \big|_{a_j'=\mu'^c_j(o_j)}\\
\end{split}
\end{equation}
where $\mu'^c = \{\mu_{\theta'^c_1}, \dots, \mu_{\theta'^c_N} \}$ is the set of target policies with delayed parameters $\theta'^c_i$.

Here, the key challenge is that policies trained by MADDPG may converge to different local optima. Therefore, the other agents may choose policies that are different from the ones learned by MADDPG. To address this, we propose to train a collection of $K$ different policies for each agent in a single scenario. Each policy can have different initial parameters and selection of the partners' policies. This will grow the populations in the policy set of each agent and further improve the robustness during testing. Unlike MADDPG, we do not ensemble the $K$ policies to a single policy but keep all the individual policies as candidates for execution.

\subsection{Learning Policy Predictors}

We denote $\phi_i: \mathcal{H}_i \rightarrow \Delta(\Pi_i)$ the policy predictor that uses agent $i$'s local observation history $h_i^t = (o_i^1, \dots, o_i^t)$ to compute the distribution over agent $i$'s policy set $\Pi_i$. Our goal is to determine at execution time which policy should be used by agent $i$ in order to achieve the best performance. Here, we use a recurrent neural network to train a policy predictor $\phi_i$, containing a layer of LSTM and some other layers. This structure allows the agent to reason about the current scenario using its observation sequence.

Here, $\phi_i(o^t_i,h^{t-1}_i)$ is a function that takes the input of the current observation $o^t_i$ and the last-step history $h^{t-1}_i$ at the time step $t$, and outputs the policy distribution $p^t_i (\cdot) \in [0, 1]$ over agent $i$'s policy set $\Pi_i$. Now, the action selection process of agent $i$ at time step $t$ can be written as:
\begin{equation}
\begin{split}
    p_i^t &= \phi_i(o_{i}^{t},h_{i}^{t-1}) \\
    \mu_i & = {\arg\max}_{\mu'_i \in \Pi_i} p^t_i(\mu'_i) \\
    a_i^t &= \mu_{\theta_i}(o_i^t)
\end{split}
\end{equation}

Together with training the policy, we use replay buffer to train $\phi_i$ in order to avoid the early instability and adverse effects during training process. Specifically, we create a dedicated replay buffer $\mathcal{B}_{i}$ for $\phi_i$ during training. It stores $(h_i, \mu_i)$ at the end of each episode, where $h_i = (o_i^1,\ldots,o_i^T)$ is agent $i$'s observation sequence at this episode and $\mu_i$ is the currently trained policy. The main training procedure of $\phi_i$ is to sample a random minibatch of samples $(h_i, \mu_i)$ from $\mathcal{B}_{i}$ and update the parameters of $\phi_i$ by minimizing the cross-entropy loss function as follow:
\begin{equation}
\begin{split}
\nabla_{p_i}J(\phi_i) & = \mathbb{E}_{(h_i, \mu_i) \sim \mathcal{B}_i}\left[\sum_{t=1}^T \mathrm{CE} \left( \phi_i(o_i^t,h_i^{t-1}),t \right) \right] \\
& = \mathbb{E}_{(h_i, \mu_i)}\left[\sum_{t=1}^T \sum_{\mu'_i \in \Pi_i} -y^{\mu'_i} \log\left( p_i^t(\mu'_i) \right) \right]\\
\mathrm{where}~ y^{\mu'_i} & =
\begin{cases}
1, & \mu'_i = \mu_i\\
0, & \mu'_i \ne \mu_i
\end{cases}
\mathrm{and}~ p_i^t = \phi_i(o_i^t, h_i^{t-1}).
\end{split}
\end{equation}

The overall learning procedures of our PAMADDPG method are outlined in Algorithm \ref{alg:learn}.

\begin{algorithm}[t]
    \caption{Learning agents' policies and predictors}
    \label{alg:learn}
    \label{algorithm_train}
    \ForEach{episode} {
        Initialize a random process $\mathcal{N}$ for action exploration \\
        Receive initial observations $\mathbf{x} = (o_{1}, \dots, o_{N})$ \\
        \For{time step $t=1$ \KwTo $T$} {
            For each agent $i$, select $a_{i} = \mu_{\theta_{i}}(o_i) + \mathcal{N}_t$ w.r.t the current policy and exploration noise \\
            Execute action $a = (a_{1},\ldots,a_{N})$ and observe reward $r = (r_{1},\ldots,r_{N})$ and new state $\mathbf{x}'$ \\
            Store $(\mathbf{x}, a, r, \mathbf{x}')$ in $\mathcal{D}$ and set $\mathbf{x} \gets \mathbf{x}'$ \\
            \For{agent $i = 1$ \KwTo $N$}{
                Sample a random minibatch of $M$ samples $(\mathbf{x}^m, a^m, r^m, \mathbf{x}'^m)$ from replay buffer $\mathcal{D}$ \\
                Set $y^m = r_i^m + \gamma\, Q^{\mu'}_i(\mathbf{x}', a') \big|_{a_{j}'=\mu'_{j}(o_j^m)}$ \\
                Update critic by minimizing the loss:
                \[\small
                \begin{split}
                \mathcal{L}(\theta_i) = \frac{1}{M} \sum_{m=1}^M \left( y^m-Q_i^{\mu}(\mathbf{x}^m,a^m) \right)^2
                \end{split}
                \] \\
                Update actor using the sampled gradient:
                \[\small
                \begin{split}
                \nabla_{\theta_i}J(\mu_i) \approx & \frac{1}{M} \sum_{m=1}^M \nabla_{\theta_i} \mu_i(o_i^m) \\ & \nabla_{a_i} Q_i^{\mu}(\mathbf{x}^m, a^m) \big|_{a_i=\mu_i(o_i^m)}
                \end{split}
                \] \\
                Sample a random minibatch of $K$ samples $(h_i^k, \mu_i^k)$ from replay buffer $\mathcal{B}_{i}$ \\
                Update predictor $\phi_i$ by minimizing the loss:
                \[\small
                \begin{split}
                \nabla_{p_i}J(\phi_i) \approx \frac{1}{K} \sum_{k=1}^K \sum_{t=1}^T \sum_{\mu'_i} -y^{\mu'_i} \log(p_i^t(\mu'_i))
                \end{split}
                \]
            }
            Update target network parameters $\theta_i$ for each agent $i$ as:
            $\theta'_i \gets \tau \theta_i + (1-\tau) \theta'_i$
        }
        Collect history $h_i = (o_i^1, \dots, o_i^T)$ and store $(h_i, \mu_i)$ in replay buffer $\mathcal{B}_{i}$ for each agent $i$
    }
\end{algorithm}


\section{Experiments}

We empirically evaluate our algorithm on three domains built on top of the particle-world environments\footnote{\scriptsize Code from: \url{https://github.com/openai/multiagent-particle-envs}} originally used by the MADDPG paper \cite{lowe2017multi}. To create various scenarios, we modify some of the physical properties of the environments so that the agents must alter their policies in order to success in different scenarios. By doing so, we expect to examine the adaptiveness of our PAMADDPG algorithm when testing in different scenarios.

\subsection{Environments}

\begin{figure}[t]
    \centering

    \subfigure[Keep-away]{
        \includegraphics[width=.3\linewidth]{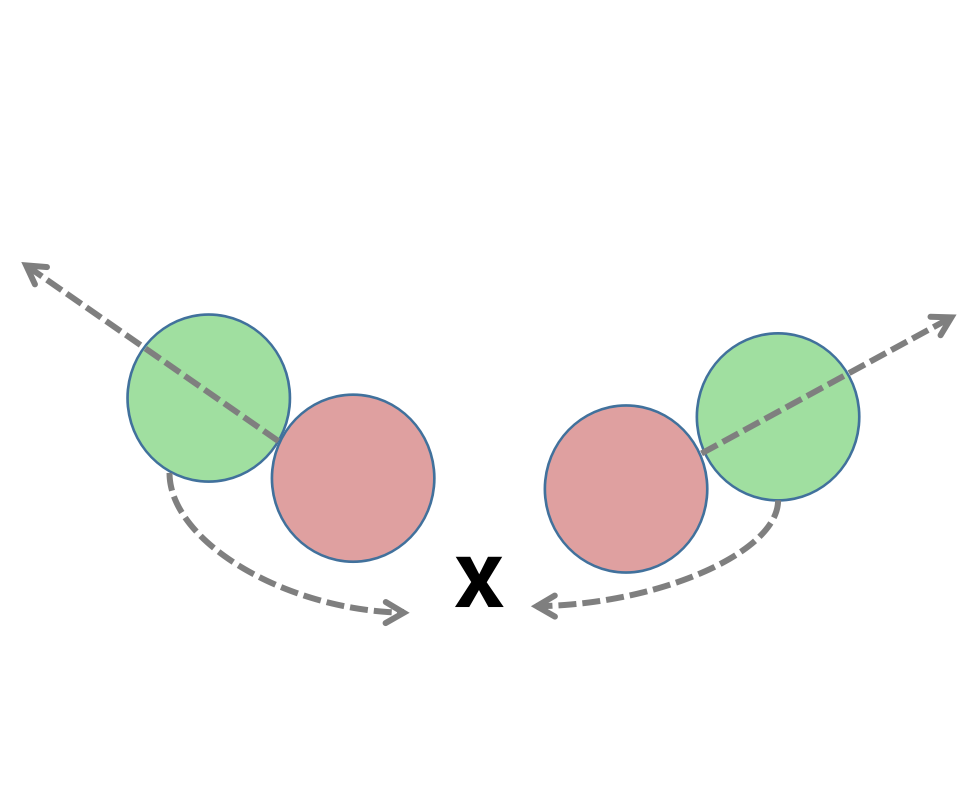}
        \label{fig:env:push}
    }
    \subfigure[Predator-prey]{
        \includegraphics[width=.3\linewidth]{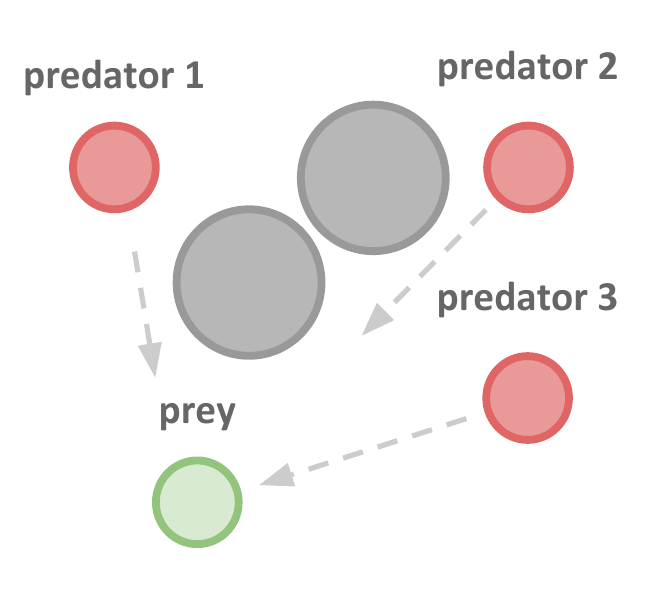}
        \label{fig:env:tag}
    }
    \subfigure[Cooperative navigation]{
        \includegraphics[width=.3\linewidth]{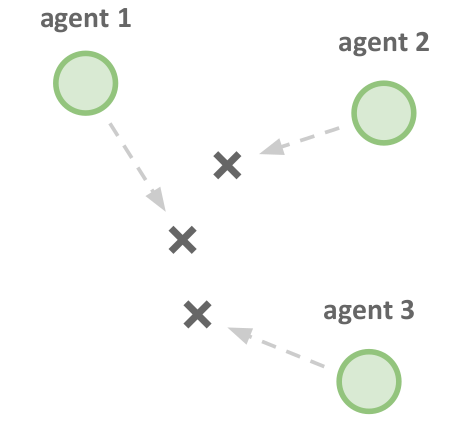}
        \label{fig:env:spread}
    }
    \caption{Illustrations of the three environments.}
    \label{fig:env:all}
\end{figure}

The particle world environment consists of $N$ cooperative agents, $M$ adversarial agents and $L$ landmarks in a two-dimensional world with continuous space. In the experiments, we consider two mixed cooperative and competitive domains (i.e., Keep-away and Predator-prey) and one fully cooperative domain (i.e., Cooperative navigation), as shown in Figure \ref{fig:env:all}, and modify these environments to generate different scenarios as below.

\paragraph{Keep-away.}
This environment consists of $L$ landmarks including a target landmark, $N = 2$ cooperating agents who know the target landmark and are rewarded based on their distance to the target, and $M = 2$ agents who must prevent the cooperating agents from reaching the target.  Adversaries accomplish this by physically pushing the agents away from the landmark, temporarily occupying it. While the adversaries are also rewarded based on their distance to the target landmark, they do not know the correct target.

We create $K=3$ scenarios that require agents to learn to adapt with. In each scenario, we simulate different ``wind'' conditions in the environment. The wind will affect the moving speed of the agents in a certain direction computed as: $v'_i = v_i + w * \beta_i$, where $v_i$ is the original speed, $w = [w_N, w_W, w_S, w_E]$ is the wind force for four directions, and $\beta_i = 5$ is the acceleration rate. In the experiments, we consider no wind (i.e., $w = 0$) in Scenario 1, southwest wind (i.e., $w_S = w_W = 0.5$ and 0 otherwise) in Scenario 2, and northeast wind (i.e., $w_N = w_E = 0.5$ and 0 otherwise) in Scenario 3 respectively.

\paragraph{Predator-prey.}
In this environment, $N=4$ slower cooperating agents must chase $M=2$ faster adversary around a randomly generated environment with $L=2$ large landmarks impeding the way. Each time the cooperative agents collide with an adversary, the agents are rewarded while the adversary is penalized. Agents observe the relative positions and velocities of the agents, and the landmark positions.

We create $K=3$ scenarios to simulate different body conditions for the good and bad agents. This is done by using different maximum speeds $\bar{v}$ and accelerations $\beta$ for the agents in the environment, i.e., $( \bar{v}_{good}, \beta_{good}, \bar{v}_{bad}, \beta_{bad} )$. We set the parameters so that the agents will compete in different levels, i.e., weak, medium, and strong. Specifically, we set $(3.0, 3.0, 3.9, 4.0)$ in Scenario 1, $(2.0, 4.0, 2.6, 5.0)$ in Scenario 2, and $(3.0, 5.0, 3.9, 6.0)$ in Scenario 3.

\paragraph{Cooperative navigation.}
In this environment, agents must cooperate through physical actions to reach a set of $L$ landmarks. Agents observe the relative positions of other agents and landmarks, and are collectively rewarded based on the proximity of any agent to each landmark. In other words, the agents have to ``cover'' all of the landmarks. Furthermore, the agents occupy significant physical space and are penalized when colliding with each other.

Similar to the Keep-away environment described above, we created $K=3$ scenarios in this environment also with three wind conditions, i.e., no wind for Scenario 1, southeast wind for Scenario 2, and northwest wind for Scenario 3.

\subsection{Setup}

We compared our PAMADDPG algorithm with MADDPG\footnote{Code from: \url{https://github.com/openai/maddpg}} and M3DDPG\footnote{Code from: \url{https://github.com/dadadidodi/m3ddpg}}, which are currently the leading algorithms for multi-agent deep RL, on the environments as described above. In our implementation, the agents' policies are represented by a two-layer ReLU MLP with 64 units per layer, which is the same as MADDPG and M3DDPG, and the policy predictors are represented by a two-layer ReLU MLP and a layer of LSTM on top of them.

We used the same training configurations as MADDPG and M3DDPG, and ran all the algorithms until convergence. Then, we tested the policies computed by the algorithms on each environment with 10,000 further episodes and report the averaged results. For fair comparison, all algorithms were tested on a fixed set of environment configurations. Each testing environment is generated by randomizing the basic configurations and randomly selecting a scenario. As aforementioned, the agents do not know which scenario is selected for the environment during testing procedure.

Note that MADDPG and M3DDPG do not consider different scenarios in their original implementations. For fair comparison, we try to train their policies in a way that their performance is improved when working with different scenarios. Specifically, in our experiments, MADDPG trained policies with all scenarios and optimized the objective as:
\begin{equation}
  J(\theta_i) = \mathbb{E}_{c \sim \mathrm{uniform}(\mathcal{C}), s \sim p^c, a \sim \mu} [ R_i(s, a) ]
\end{equation}
As aforementioned, we do not know the true distribution before testing so MADDPG was trained with the uniformly distributed scenarios. Following the minmax idea of the standard version, M3DDPG maximized the objective in the worst-case scenario in the experiments as:
\begin{equation}
  J(\theta_i) = \min\nolimits_{c \in \mathcal{C}, a_{j\neq i}} \mathbb{E}_{s \sim p^c, a_i \sim \mu_i} [ R_i(s, a) ]
\end{equation}
By doing so, we can evaluate the effectiveness of our algorithm with multiple policies comparing with MADDPG and M3DDPG using only a single policy for each agent when the environment changes.

\subsection{Results}

\begin{figure}[t]
    \centering
    \includegraphics[width=\linewidth]{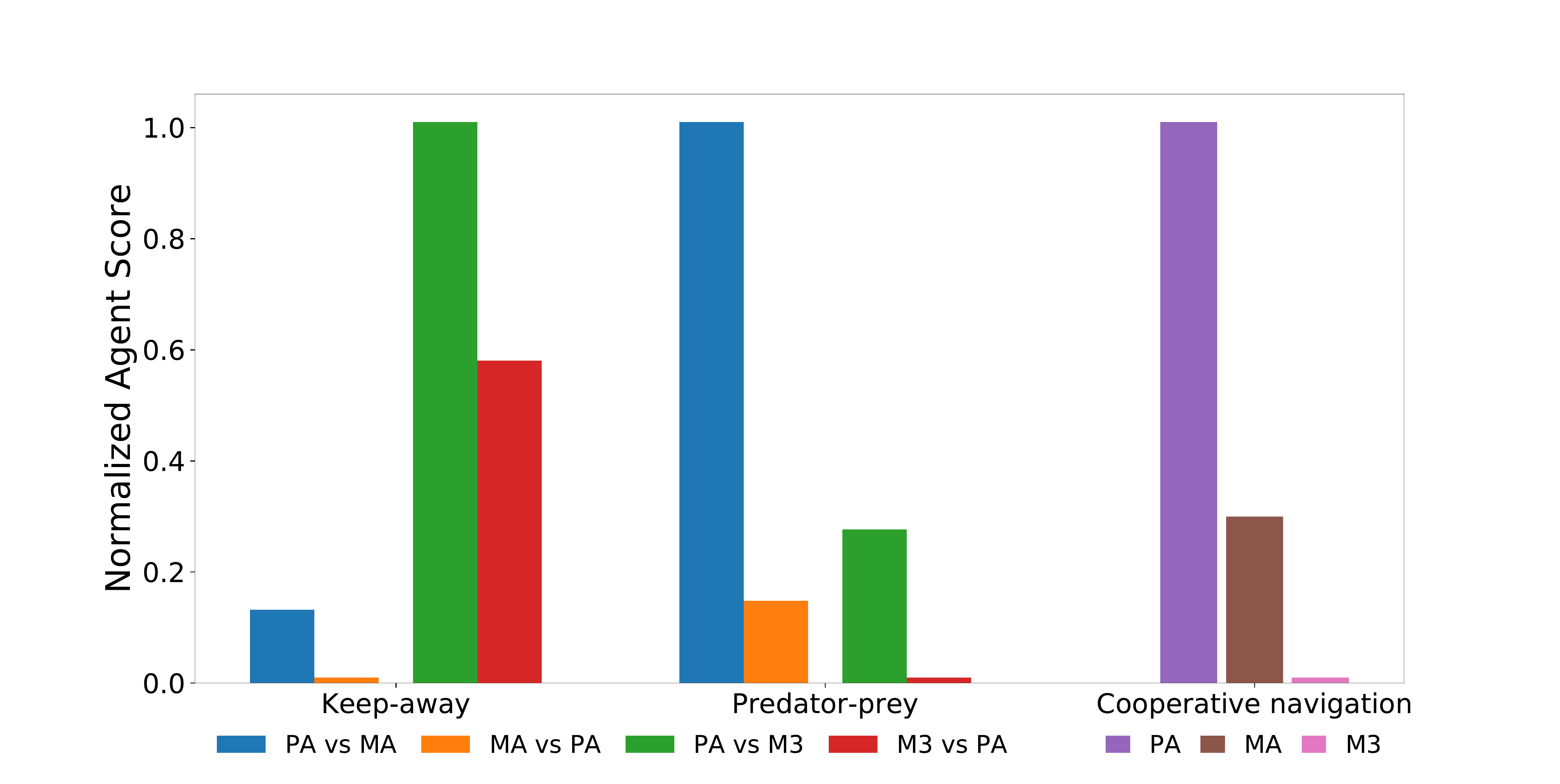}
    \caption{Overall performance of PAMADDPG (PA), MADDPG (MA), and M3DDPG (M3) on the environments.}
    \label{fig:bar:total}
\end{figure}

\begin{figure}[t]
    \centering
    \subfigure[Keep-away]{
        \includegraphics[width=\linewidth]{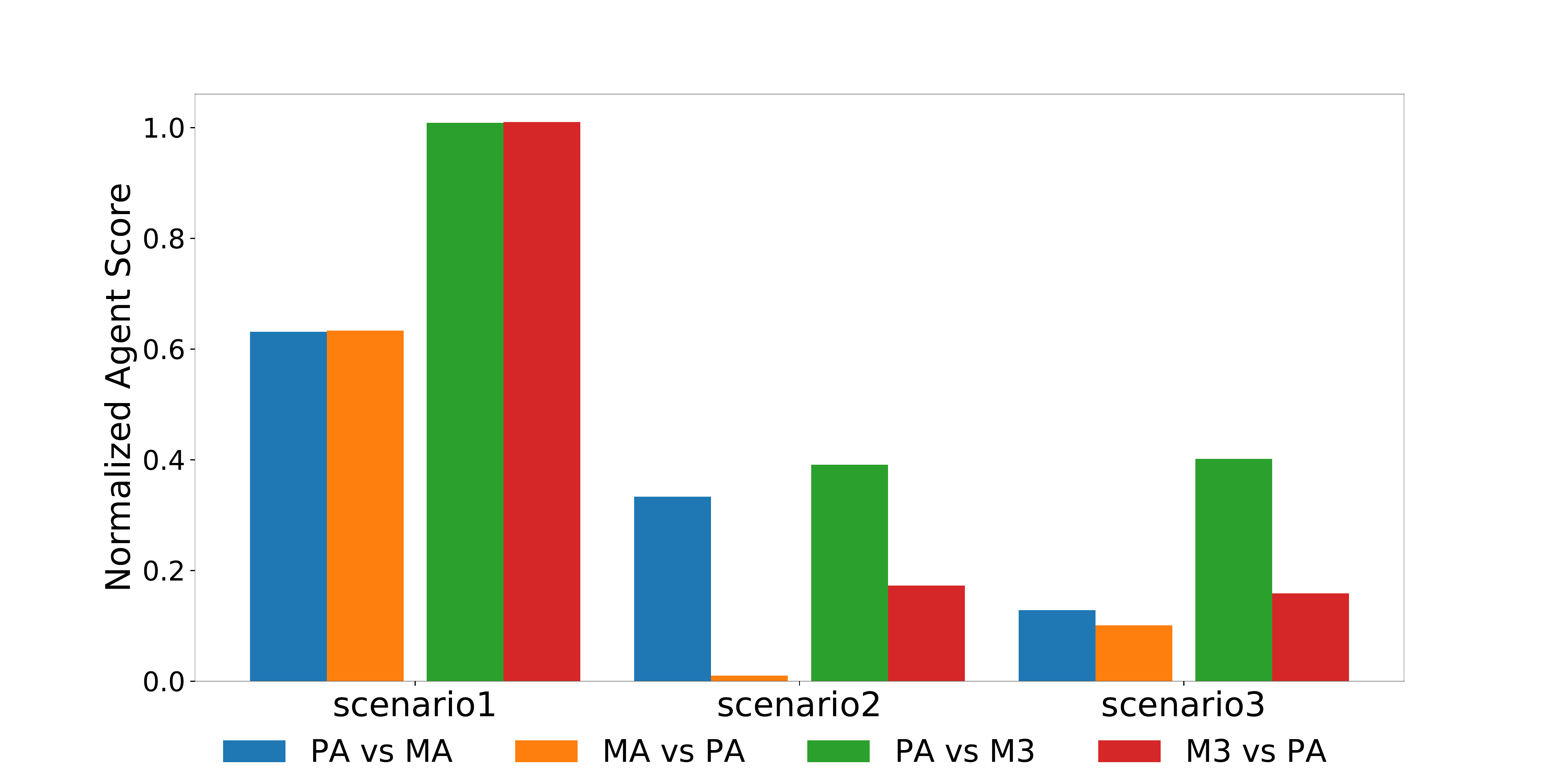}
        \label{fig:bar:push}
    }
    \subfigure[Predator-prey]{
        \includegraphics[width=\linewidth]{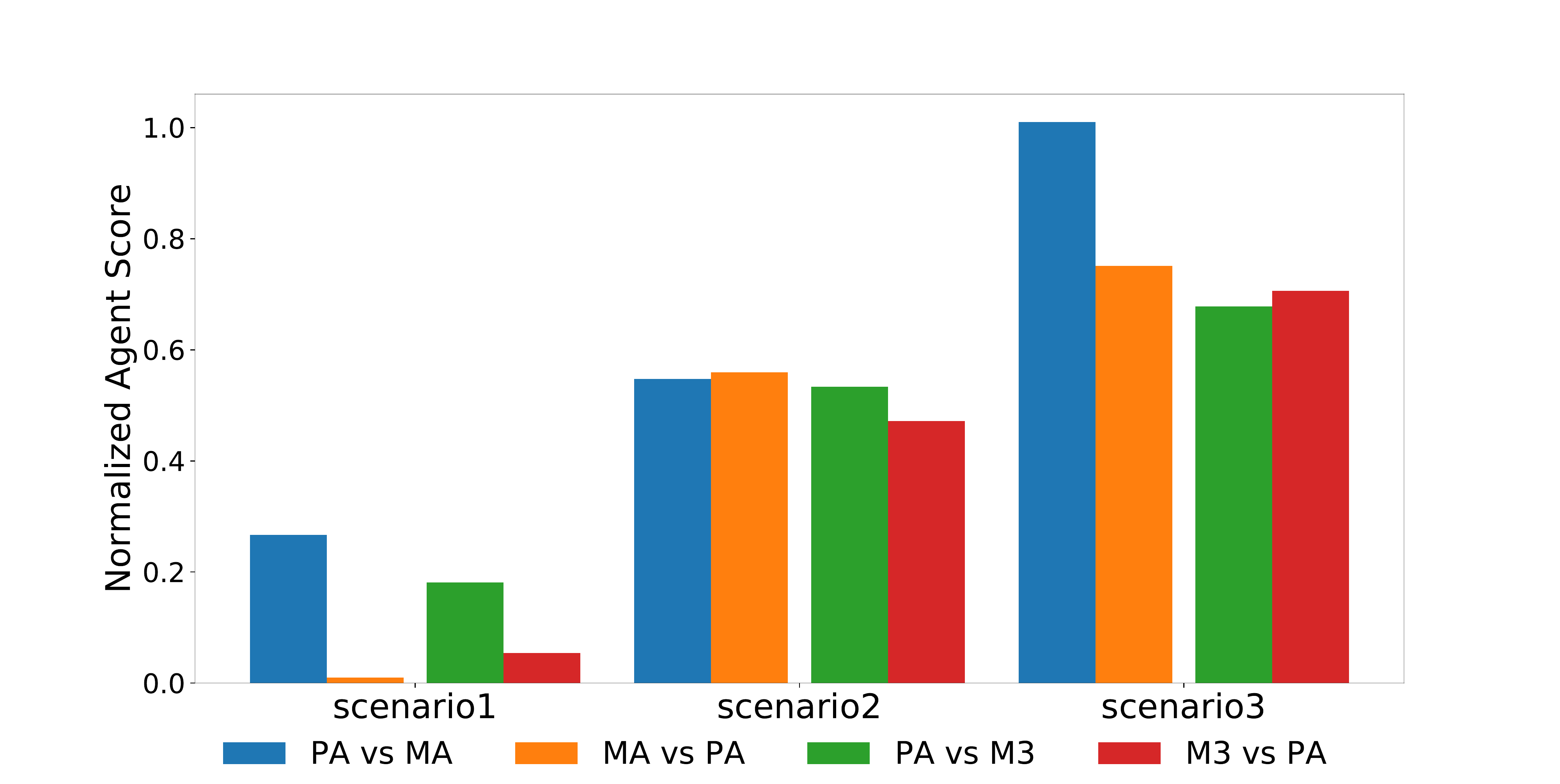}
        \label{fig:bar:tag}
    }
    \subfigure[Cooperative navigation]{
        \includegraphics[width=\linewidth]{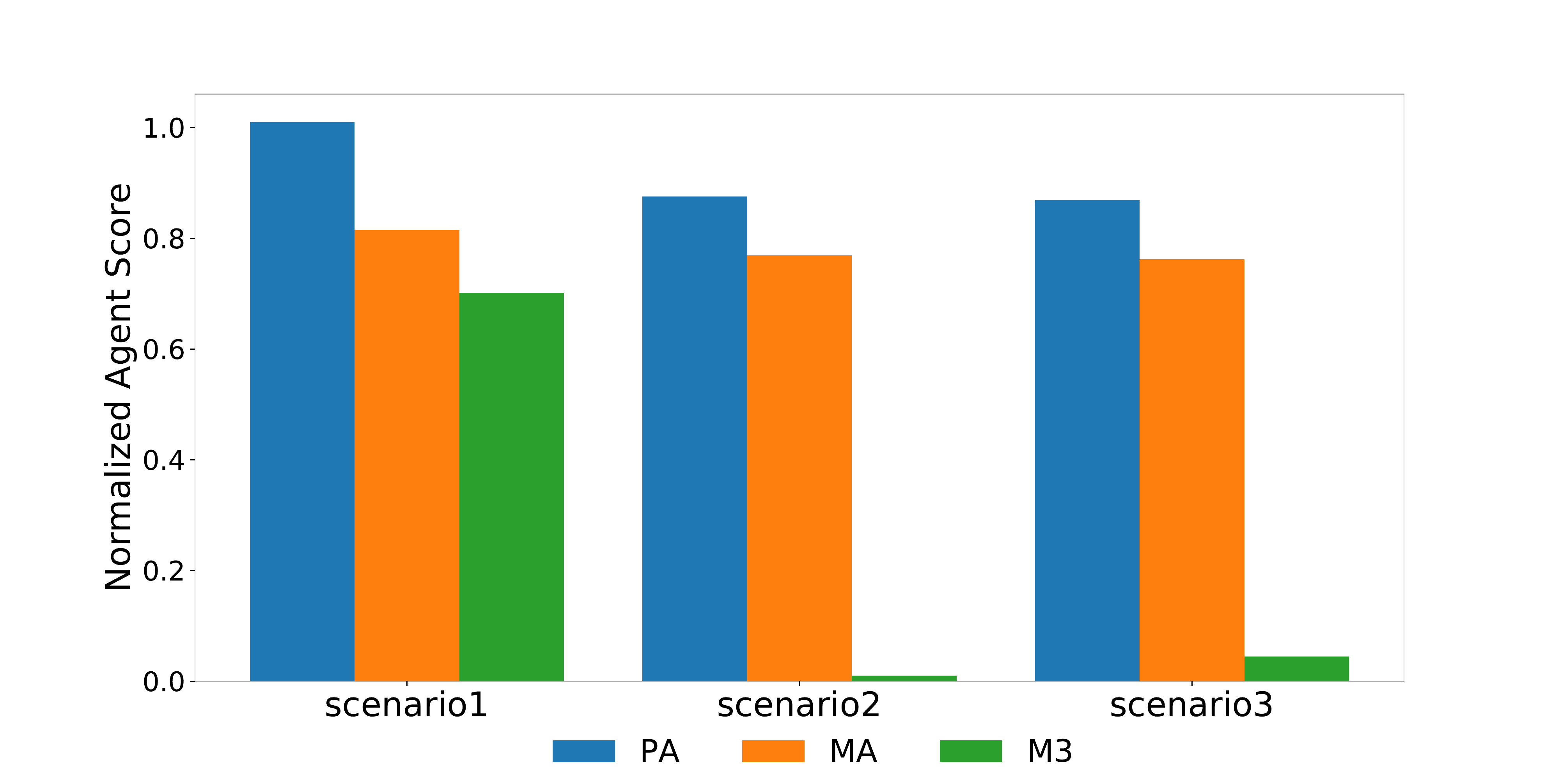}
        \label{fig:bar:spread}
    }
    \caption{Performance of PAMADDPG (PA), MADDPG (MA), and M3DDPG (M3) on different scenarios.}
    \label{fig:bar:scenarios}
\end{figure}

\begin{figure}[t]
    \centering
    \includegraphics[width=\linewidth]{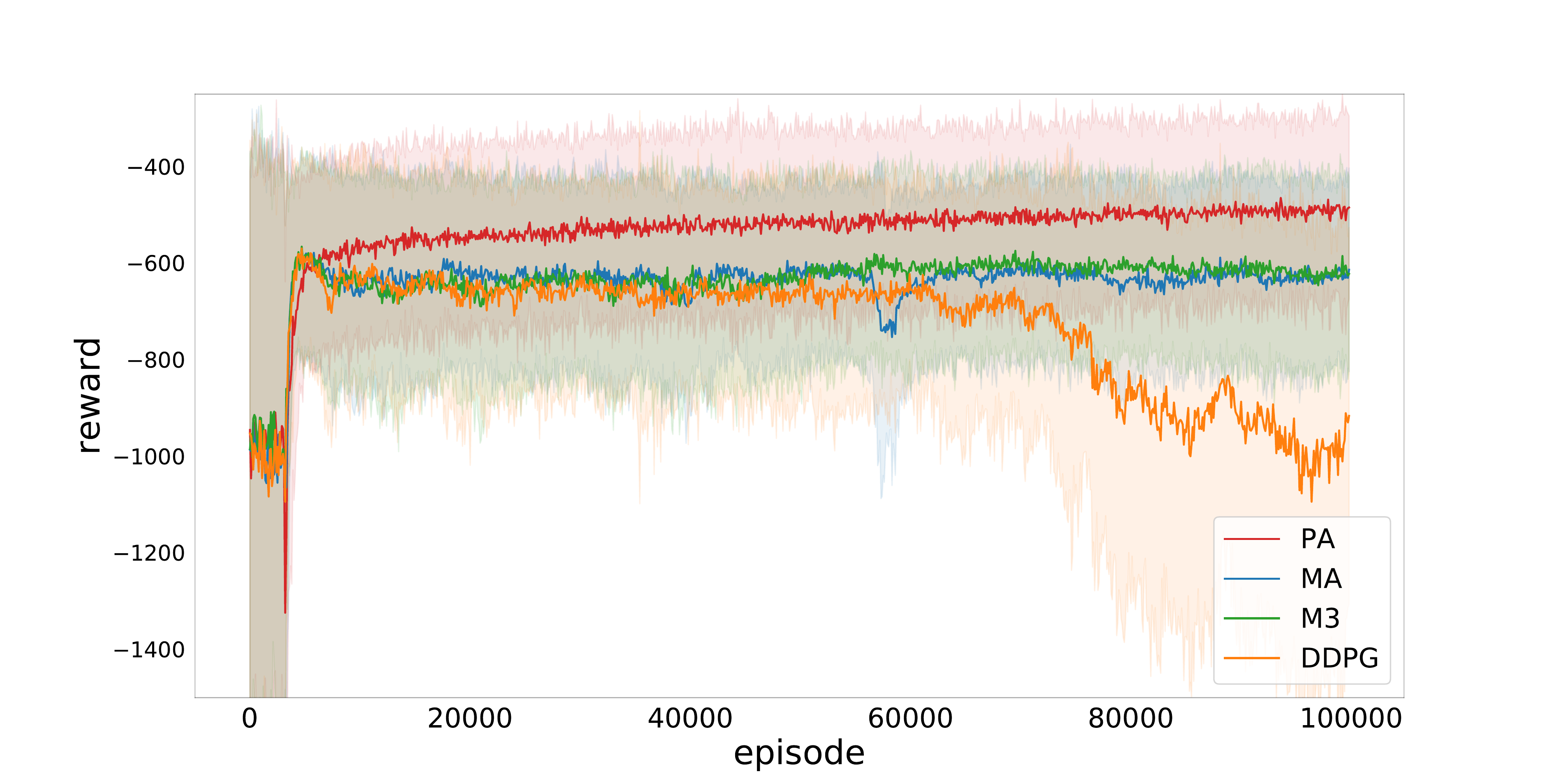}
    \caption{Learning reward of PAMADDPG (PA), MADDPG (MA), M3DDPG (M3), and DDPG on the Cooperative navigation environment after 10,000 episodes.}
    \label{fig:train:spread}
\end{figure}

We measure the performance of agents with policies learned by our PAMADDPG and agents with policies learned by MADDPG and M3DDPG in each environment. In the first two mixed cooperative and competitive domains, we switch the roles of both normal agent and adversary as in the MADDPG and M3DDPG papers to evaluate the quality of learned policies trained by different algorithms.

The results on the three environments are demonstrated in Figure \ref{fig:bar:total}. As shown in the figure, each group of bar shows the 0$-$1 normalized score for the environment, where a higher score shows better performance for the algorithm. In the first two environments, PAMADDPG outperforms M3DDPG and MADDPG because PAMADDPG achieves higher scores when playing normal agents (i.e., PA vs MA, PA vs M3) than the ones as adversaries (i.e., MA vs PA, M3 vs PA). Interestingly, PAMADDPG performs better when playing against MADDPG (i.e., PA vs MA, MA vs PA) than the case against M3DDPG (i.e., PA vs M3, M3 vs PA) in the Keep-away environment, while PAMADDPG shows better performance against M3DDPG than the case against MADDPG in the Predator-prey environment. Intuitively, this is because the Predator-prey environment is more competitive than the Keep-away environment so that M3DDPG who considers the worst-case situation works better than MADDPG when paired with our algorithm. In the Cooperative navigation environment, PAMADDPG consistently outperforms MADDPG and M3DDPG. M3DDPG has the worst performance in terms of scores because this environment is a fully cooperative domain while M3DDPG makes unrealistic assumption that all the other agents act adversarially.

Figure \ref{fig:bar:scenarios} shows the results of our PAMADDPG comparing with MADDPG and M3DDPG when testing on different scenarios in each environment. In the Keep-away environment, PAMADDPG outperforms MADDPG and M3DDPG on Scenarios 2 and 3 while performs similarly on Scenario 1. This is because MADDPG and M3DDPG tends to converge to the policies fitting Scenario 1, which is expected to work poorly in Scenarios 2 and 3. In contrast, our PAMADDPG can adapt its policies to fit different scenarios during testing. In the Predator-prey environment, PAMADDPG outperforms MADDPG on Scenarios 1 and 3 but not Scenario 2, and M3DDPG on Scenarios 1 and 2 but not Scenario 3. Similar to the Keep-away environment, this is because MADDPG converges to the policies fitting Scenario 2 while M3DDPG converges to the policies fitting Scenario 3. As we can see from the figure, PAMADDPG achieves slightly less scores than MADDPG and M3DDPG on Scenarios 2 and 3 respectively. This is because the Predator-prey environment is very competitive and the policy predictors in PAMADDPG take time to form correct predictions. In the Cooperative navigation environment, our PAMADDPG outperforms MADDPG and M3DDPG for all the scenarios. Again, M3DDPG has the worst performance because this is a fully cooperative environment.

Figure \ref{fig:train:spread} shows the average reward of different approaches on the Cooperative navigation environment during the training process. As we can see from the figure, our PAMADDPG algorithm converges to better reward than all the other methods. As expected, the reward of DDPG decreases after 80,000 episodes due to non-stationarity in multi-agent RL. As shown in the figure, the reward of MADDPG fluctuates about 60,000 episodes while the reward of our PAMADDPG becomes stable after convergence.

\section{Conclusion}

In this paper, we addressed the non-stationarity problem in multi-agent RL and proposed the PAMADDPG algorithm. we model the non-stationarity in the environment as a finite set of scenarios. At training time, each agent learns multiple policies, one for each scenario, and trains a policy predictor that can be used to predict the best policy during execution. With the multiple policies and policy predictor, each agent is able to adapt its policy and choose the best one for the current scenario. We tested our algorithm on three common benchmark environments and showed that PAMADDPG outperforms MADDPG and M3DDPG in all the tested environment. In the future, we plan to conduct research on learning the scenarios directly from the environment.


\bibliographystyle{aaai}
{\small \bibliography{citations}}

\begin{thebibliography}{}

\bibitem[\protect\citeauthoryear{Al-Shedivat \bgroup et al\mbox.\egroup
  }{2017}]{al2017continuous}
Al-Shedivat, M.; Bansal, T.; Burda, Y.; Sutskever, I.; Mordatch, I.; and
  Abbeel, P.
\newblock 2017.
\newblock Continuous adaptation via meta-learning in nonstationary and
  competitive environments.
\newblock {\em arXiv preprint arXiv:1710.03641}.

\bibitem[\protect\citeauthoryear{Bansal \bgroup et al\mbox.\egroup
  }{2017}]{bansal2017emergent}
Bansal, T.; Pachocki, J.; Sidor, S.; Sutskever, I.; and Mordatch, I.
\newblock 2017.
\newblock Emergent complexity via multi-agent competition.
\newblock {\em arXiv preprint arXiv:1710.03748}.

\bibitem[\protect\citeauthoryear{Finn, Abbeel, and
  Levine}{2017}]{finn2017model}
Finn, C.; Abbeel, P.; and Levine, S.
\newblock 2017.
\newblock Model-agnostic meta-learning for fast adaptation of deep networks.
\newblock In {\em Proceedings of the 34th International Conference on Machine
  Learning},  1126--1135.

\bibitem[\protect\citeauthoryear{Foerster \bgroup et al\mbox.\egroup
  }{2016a}]{foerster2016learningto}
Foerster, J.; Assael, I.~A.; de~Freitas, N.; and Whiteson, S.
\newblock 2016a.
\newblock Learning to communicate with deep multi-agent reinforcement learning.
\newblock In {\em Proceedings of the Advances in Neural Information Processing
  Systems},  2137--2145.

\bibitem[\protect\citeauthoryear{Foerster \bgroup et al\mbox.\egroup
  }{2016b}]{foerster2016learning}
Foerster, J.~N.; Assael, Y.~M.; de~Freitas, N.; and Whiteson, S.
\newblock 2016b.
\newblock Learning to communicate to solve riddles with deep distributed
  recurrent q-networks.
\newblock {\em arXiv preprint arXiv:1602.02672}.

\bibitem[\protect\citeauthoryear{Foerster \bgroup et al\mbox.\egroup
  }{2017}]{foerster2017stabilising}
Foerster, J.; Nardelli, N.; Farquhar, G.; Afouras, T.; Torr, P.~H.; Kohli, P.;
  and Whiteson, S.
\newblock 2017.
\newblock Stabilising experience replay for deep multi-agent reinforcement
  learning.
\newblock In {\em Proceedings of the 34th International Conference on Machine
  Learning},  1146--1155.

\bibitem[\protect\citeauthoryear{Foerster \bgroup et al\mbox.\egroup
  }{2018a}]{foerster2018learning}
Foerster, J.; Chen, R.~Y.; Al-Shedivat, M.; Whiteson, S.; Abbeel, P.; and
  Mordatch, I.
\newblock 2018a.
\newblock Learning with opponent-learning awareness.
\newblock In {\em Proceedings of the 17th International Conference on
  Autonomous Agents and MultiAgent Systems},  122--130.

\bibitem[\protect\citeauthoryear{Foerster \bgroup et al\mbox.\egroup
  }{2018b}]{foerster2018counterfactual}
Foerster, J.~N.; Farquhar, G.; Afouras, T.; Nardelli, N.; and Whiteson, S.
\newblock 2018b.
\newblock Counterfactual multi-agent policy gradients.
\newblock In {\em Proceedings of the 32nd AAAI Conference on Artificial
  Intelligence}.

\bibitem[\protect\citeauthoryear{He \bgroup et al\mbox.\egroup
  }{2016}]{he2016opponent}
He, H.; Boyd-Graber, J.; Kwok, K.; and Daum{\'e}~III, H.
\newblock 2016.
\newblock Opponent modeling in deep reinforcement learning.
\newblock In {\em Proceedings of the International Conference on Machine
  Learning},  1804--1813.

\bibitem[\protect\citeauthoryear{Hernandez-Leal \bgroup et al\mbox.\egroup
  }{2017}]{hernandez2017survey}
Hernandez-Leal, P.; Kaisers, M.; Baarslag, T.; and de~Cote, E.~M.
\newblock 2017.
\newblock A survey of learning in multiagent environments: Dealing with
  non-stationarity.
\newblock {\em arXiv preprint arXiv:1707.09183}.

\bibitem[\protect\citeauthoryear{Lazaridou, Peysakhovich, and
  Baroni}{2016}]{lazaridou2016multi}
Lazaridou, A.; Peysakhovich, A.; and Baroni, M.
\newblock 2016.
\newblock Multi-agent cooperation and the emergence of (natural) language.
\newblock {\em arXiv preprint arXiv:1612.07182}.

\bibitem[\protect\citeauthoryear{Levine \bgroup et al\mbox.\egroup
  }{2016}]{levine2016end}
Levine, S.; Finn, C.; Darrell, T.; and Abbeel, P.
\newblock 2016.
\newblock End-to-end training of deep visuomotor policies.
\newblock {\em Journal of Machine Learning Research} 17(1):1334--1373.

\bibitem[\protect\citeauthoryear{Li \bgroup et al\mbox.\egroup
  }{2019}]{li2019robust}
Li, S.; Wu, Y.; Cui, X.; Dong, H.; Fang, F.; and Russell, S.
\newblock 2019.
\newblock Robust multi-agent reinforcement learning via minimax deep
  deterministic policy gradient.
\newblock In {\em Proceedings of the AAAI Conference on Artificial
  Intelligence}.

\bibitem[\protect\citeauthoryear{Lillicrap \bgroup et al\mbox.\egroup
  }{2015}]{lillicrap2015continuous}
Lillicrap, T.~P.; Hunt, J.~J.; Pritzel, A.; Heess, N.; Erez, T.; Tassa, Y.;
  Silver, D.; and Wierstra, D.
\newblock 2015.
\newblock Continuous control with deep reinforcement learning.
\newblock {\em arXiv preprint arXiv:1509.02971}.

\bibitem[\protect\citeauthoryear{Littman}{1994}]{littman1994markov}
Littman, M.~L.
\newblock 1994.
\newblock Markov games as a framework for multi-agent reinforcement learning.
\newblock In {\em Proceedings of the 11th International Conference on Machine
  Learning},  157--163.

\bibitem[\protect\citeauthoryear{Long \bgroup et al\mbox.\egroup
  }{2018}]{long2018towards}
Long, P.; Fanl, T.; Liao, X.; Liu, W.; Zhang, H.; and Pan, J.
\newblock 2018.
\newblock Towards optimally decentralized multi-robot collision avoidance via
  deep reinforcement learning.
\newblock In {\em Proceedings of the 2018 IEEE International Conference on
  Robotics and Automation},  6252--6259.

\bibitem[\protect\citeauthoryear{Lowe \bgroup et al\mbox.\egroup
  }{2017}]{lowe2017multi}
Lowe, R.; Wu, Y.; Tamar, A.; Harb, J.; Abbeel, O.~P.; and Mordatch, I.
\newblock 2017.
\newblock Multi-agent actor-critic for mixed cooperative-competitive
  environments.
\newblock In {\em Proceedings of the Advances in Neural Information Processing
  Systems},  6379--6390.

\bibitem[\protect\citeauthoryear{Mnih \bgroup et al\mbox.\egroup
  }{2015}]{mnih2015human}
Mnih, V.; Kavukcuoglu, K.; Silver, D.; Rusu, A.~A.; Veness, J.; Bellemare,
  M.~G.; Graves, A.; Riedmiller, M.; Fidjeland, A.~K.; Ostrovski, G.; et~al.
\newblock 2015.
\newblock Human-level control through deep reinforcement learning.
\newblock {\em Nature} 518(7540):529.

\bibitem[\protect\citeauthoryear{Papoudakis \bgroup et al\mbox.\egroup
  }{2019}]{papoudakis2019dealing}
Papoudakis, G.; Christianos, F.; Rahman, A.; and Albrecht, S.~V.
\newblock 2019.
\newblock Dealing with non-stationarity in multi-agent deep reinforcement
  learning.
\newblock {\em arXiv preprint arXiv:1906.04737}.

\bibitem[\protect\citeauthoryear{Raileanu \bgroup et al\mbox.\egroup
  }{2018}]{raileanu2018modeling}
Raileanu, R.; Denton, E.; Szlam, A.; and Fergus, R.
\newblock 2018.
\newblock Modeling others using oneself in multi-agent reinforcement learning.
\newblock {\em arXiv preprint arXiv:1802.09640}.

\bibitem[\protect\citeauthoryear{Silver \bgroup et al\mbox.\egroup
  }{2014}]{silver2014deterministic}
Silver, D.; Lever, G.; Heess, N.; Degris, T.; Wierstra, D.; and Riedmiller, M.
\newblock 2014.
\newblock Deterministic policy gradient algorithms.
\newblock In {\em Proceeding of the 31st International Conference on Machine
  Learning},  387--395.

\bibitem[\protect\citeauthoryear{Silver \bgroup et al\mbox.\egroup
  }{2016}]{silver2016mastering}
Silver, D.; Huang, A.; Maddison, C.~J.; Guez, A.; Sifre, L.; Van Den~Driessche,
  G.; Schrittwieser, J.; Antonoglou, I.; Panneershelvam, V.; Lanctot, M.;
  et~al.
\newblock 2016.
\newblock Mastering the game of go with deep neural networks and tree search.
\newblock {\em Nature} 529(7587):484.

\bibitem[\protect\citeauthoryear{Silver \bgroup et al\mbox.\egroup
  }{2017}]{silver2017mastering}
Silver, D.; Hubert, T.; Schrittwieser, J.; Antonoglou, I.; Lai, M.; Guez, A.;
  Lanctot, M.; Sifre, L.; Kumaran, D.; Graepel, T.; et~al.
\newblock 2017.
\newblock Mastering chess and shogi by self-play with a general reinforcement
  learning algorithm.
\newblock {\em arXiv preprint arXiv:1712.01815}.

\bibitem[\protect\citeauthoryear{Singh, Jain, and
  Sukhbaatar}{2018}]{singh2018learning}
Singh, A.; Jain, T.; and Sukhbaatar, S.
\newblock 2018.
\newblock Learning when to communicate at scale in multiagent cooperative and
  competitive tasks.
\newblock {\em arXiv preprint arXiv:1812.09755}.

\bibitem[\protect\citeauthoryear{Sukhbaatar, Fergus, and
  others}{2016}]{sukhbaatar2016learning}
Sukhbaatar, S.; Fergus, R.; et~al.
\newblock 2016.
\newblock Learning multiagent communication with backpropagation.
\newblock In {\em Proceedings of the Advances in Neural Information Processing
  Systems},  2244--2252.

\bibitem[\protect\citeauthoryear{Wu \bgroup et al\mbox.\egroup
  }{2017}]{wu2017flow}
Wu, C.; Kreidieh, A.; Parvate, K.; Vinitsky, E.; and Bayen, A.~M.
\newblock 2017.
\newblock Flow: Architecture and benchmarking for reinforcement learning in
  traffic control.
\newblock {\em arXiv preprint arXiv:1710.05465}.

\end{thebibliography}

\end{document}